\documentclass{ifacconf}

\usepackage{graphicx}      
\usepackage{natbib}        
\usepackage{amsfonts}
\usepackage{amsmath}
\usepackage{multirow}
\usepackage{xcolor}
\usepackage{booktabs,caption}
\usepackage{subcaption}
\usepackage[flushleft]{threeparttable}
\usepackage{pifont}%
\usepackage{url}
\newcommand{\cmark}{\ding{51}}%
\newcommand{\xmark}{\ding{55}}%

\begin{document}
\begin{frontmatter}

\title{Data-driven Kinematic Modeling in Soft Robots: System Identification and Uncertainty Quantification\thanksref{footnoteinfo}} 

\thanks[footnoteinfo]{This work was supported by the COALESCE: COntext Aware LEarning for Sustainable CybEr-Agricultural Systems (CPS Frontier \#1954556).}

\author[First]{Zhanhong Jiang} 
\author[Second]{Dylan Shah}
\author[First]{Hsin-Jung Yang}
\author[First]{Soumik Sarkar}

\address[First]{Iowa State University, 
   Ames, IA 50011 USA (e-mail: zhjiang@iastate.edu,
   hjy@iastate.edu,
   soumiks@iastate.edu).}
\address[Second]{Arieca, Inc., Pittsburgh, PA 15208 USA 
    (e-mail: dshah@arieca.com)}

\begin{abstract}                
Precise kinematic modeling is critical in calibration and controller design for soft robots, yet remains a challenging issue due to their highly nonlinear and complex behaviors. To tackle the issue, numerous data-driven machine learning approaches have been proposed for modeling nonlinear dynamics. However, these models suffer from prediction uncertainty that can negatively affect modeling accuracy, and uncertainty quantification for kinematic modeling in soft robots is underexplored. In this work, using limited simulation and real-world data, we first investigate multiple linear and nonlinear machine learning models commonly used for kinematic modeling of soft robots. The results reveal that nonlinear ensemble methods exhibit the most robust generalization performance. We then develop a conformal kinematic modeling framework for soft robots by utilizing split conformal prediction to quantify predictive position uncertainty, ensuring distribution-free prediction intervals with a theoretical guarantee. 

\end{abstract}


\end{frontmatter}

\section{Introduction}
\vspace{-0.2in}
Soft robots have been shown to outperform rigid robots with high flexibility, compliance, and adaptability in complex surrounding environments~\citep{pinskier2022bioinspiration}, which has motivated a broad range of research topics ranging from control strategies to bioinspired design, to further improve their utility~\citep{pinskier2022bioinspiration}. Emerging applications of soft robots include soft grippers for handling fragile objects~\citep{zhou2021bio}, mechanoreceptive sensing using soft sensors~\citep{seo2024soft}, intelligent robotic perception~\citep{wang2021highly}, and safe human-robot interaction~\citep{wang2024perceived}. Despite these successes, establishing accurate soft robotic kinematic models remains a challenging topic, primarily due to the complex yet unpredictable behaviors that stem from structural compliance and viscoelasticity in the material~\citep{dou2021soft}. These behaviors naturally lead to highly nonlinear relationships between the system input and output, demanding effective modeling techniques.

To attenuate the above issue, diverse modeling techniques
have been developed utilizing advances from various disciplines, such as continuum mechanics, geometrical models, and discrete models~\citep{armanini2023soft}. Though they can describe underlying kinematics properly, a substantial amount of physics- and geometric-based knowledge is required, posing difficulties in fast and accurate modeling. Machine learning (ML) models have recently attracted considerable attention, as they are well-known in solving nonlinear modeling problems. Relevant applications in soft robots include soft sensor calibration and positioning control of soft actuators~\citep{abbasi2020position}, grasping~\citep{arapi2020grasp}, and motion planning~\citep{jitosho2023reinforcement}. Notably, deep learning (DL) models can also advance modeling and control~\citep{beaber2024physics} in this area, but a large amount of data is required. In addition to this, ML models lack interpretability, which until now has made uncertainty quantification difficult.

To bridge the above gaps, in this work, we first conduct an extensive study on data-driven kinematic modeling by using multiple linear and nonlinear ML methods, and apply the models to both simulation and real-world datasets. One key finding is that nonlinear ensemble methods are performant and generalizable in predictive positional modeling for soft robots. To ensure reliable model prediction results, we develop \textit{conformal kinematic modeling} by resorting to conformal prediction approaches on selected nonlinear ensemble methods such that the modeling uncertainty can be quantified. The developed conformal kinematic modeling method provably provides a guaranteed coverage of the true positions of a soft robot within a specified prediction set, ensuring a certain level of confidence in the predictions. This lays the foundation for more efficient stochastic optimization and controller design for soft robots.
\section{Problem Formulation}
\vspace{-0.1in}
In this work, we consider a soft robot actuated with $n$ different actuation commands denoted by $\mathbf{u}\in\mathbb{R}^n$ and with $m$ end effectors. We denote by $\mathbf{x}\in\mathbb{R}^{3m}$ the true position of $m$ end effectors. The forward kinematics of the soft robot is denoted as $\mathcal{K}_f$, mapping the actuation commands $\mathbf{u}$ to estimated positions of end effectors denoted as $\hat{\mathbf{x}}$. Throughout the work, kinematics parameters are denoted as $\theta\in\mathbb{R}^d$ such that the forward kinematics is represented by the following formula:
\begin{equation}\label{eq_1}
    \hat{\mathbf{x}} = \mathcal{K}_f(\mathbf{u};\theta).
\end{equation}
If $\mathcal{K}_f$ is described by physics-based knowledge and the dimension of $\theta$ is relatively low, the robot state can be known with high accuracy. However, this requires practitioners to gain such complicated knowledge a priori, which is often not possible, especially with complicated robots. Data-driven techniques may provide a way to learn $\mathcal{K}_f$ in a more generalized sense. 
Kinematic models of soft robots involve structural and joint type parameters which are typically defined separately in data-driven methods. In this work, we do not explicitly define them in $\mathcal{K}_f$, but instead lump them together as $\theta$. Thus, given the data of actuation commands and positions $\{(\mathbf{u}, \mathbf{x})\}$, we can train data-driven methods to approximate $\mathcal{K}_f$.

\begin{figure}
    \centering
    \includegraphics[width=0.9\linewidth]{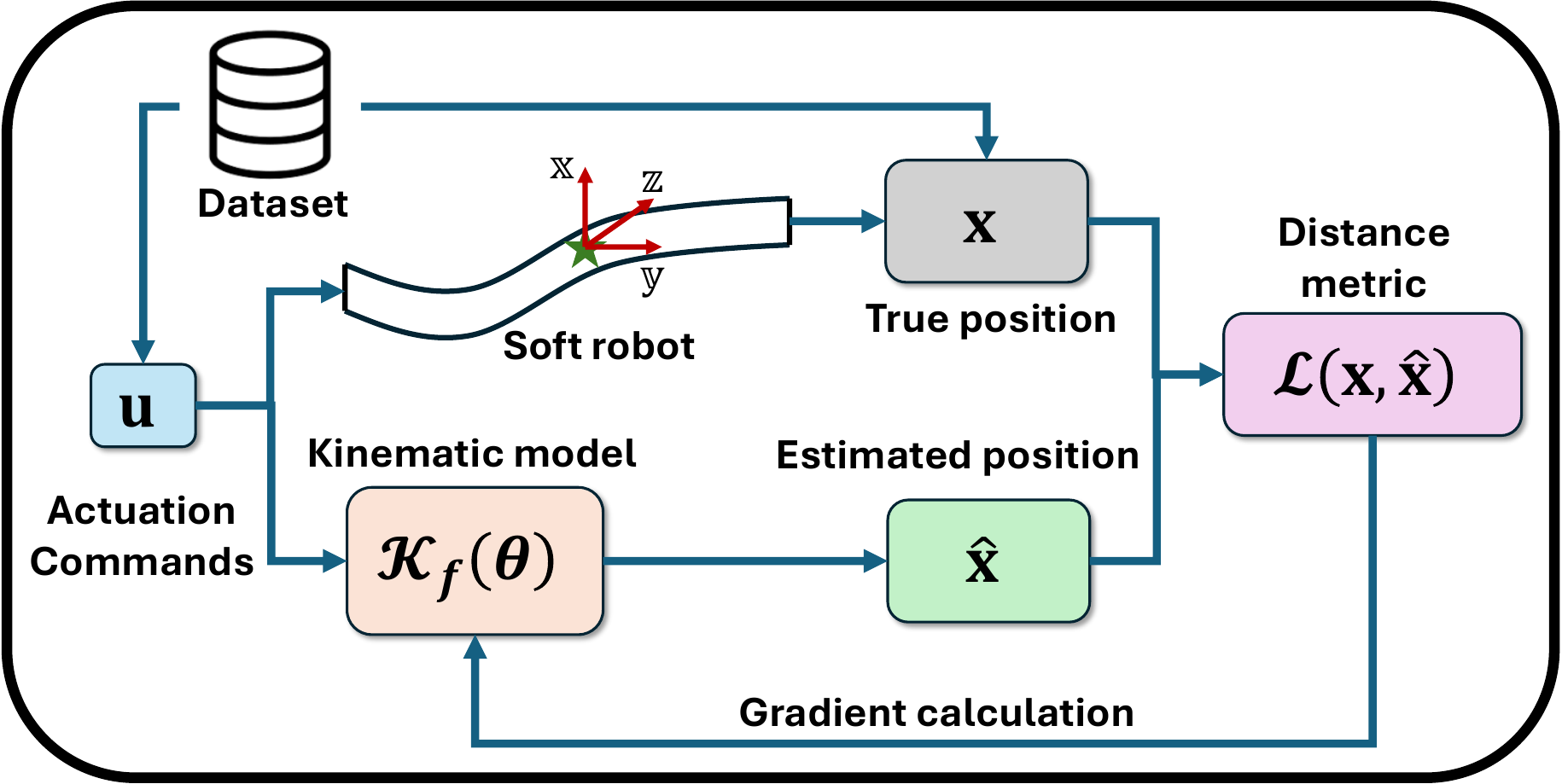}
    \caption{Schematic diagram of system identification of soft robots.}
    \label{fig:si_sr}
\end{figure}

Next, we conduct system identification process, searching for the optimal parameters $\theta$ in $\mathcal{K}_f$ to minimize an objective function, such as a distance metric $\mathcal{L}(\mathbf{x}, \hat{\mathbf{x}})$ between the true position $\mathbf{x}$ and the estimated position $\hat{\mathbf{x}}$:
\begin{equation}\label{eq_2}
    \hat{\theta} = \text{argmin}_{\theta\in\mathbb{R}^d}\mathbb{E}_{(\mathbf{u},\mathbf{x})\in\mathcal{D}}[\mathcal{L}(\mathbf{x}, \mathcal{K}_f(\mathbf{u};\theta))],
\end{equation}
where $\mathcal{D}$ is the collected dataset of actuation commands and groundtruth positions (Figure~\ref{fig:si_sr}). Eq.~\ref{eq_2} is the so-called \textit{expected risk minimization}, where the distribution of $\mathcal{D}$ is unknown, and practically infeasible to be solved. Thus, we convert it to the popular \textit{empirical risk minimization}:
\begin{equation}\label{eq_3}
        \hat{\theta} = \text{argmin}_{\theta\in\mathbb{R}^d}\frac{1}{N}\sum_{(\mathbf{u},\mathbf{x})\in\mathcal{D}}\mathcal{L}(\mathbf{x}, \mathcal{K}_f(\mathbf{u};\theta)),
\end{equation}
where $N$ is the size of dataset $\mathcal{D}$. We can then adopt gradient descent type of algorithms such as SGD or Adam to solve Eq.~\ref{eq_3}. 
This is a regression task such that the popular mean absolute error (MAE) or mean squared error (MSE) is used to quantify $\mathcal{L}$. The complexity of $\mathcal{K}_f$ can cause difficulties in finding the globally optimal solution. Thus, we apply different categories of ML approaches, to provide insights such as the impact of model complexity on the regression performance when modeling soft robots.

\section{Methods}
\vspace{-0.1in}

In this study, we employ four main classes of approaches (Table~\ref{tb:ML}).  
Predictions generated by ML models suffer from uncertainty, which degrades the reliability of the result. To quantify this uncertainty, here we introduce a technique called conformal prediction (CP)~\citep{romano2019conformalized}, and apply it to the studied ML regression models to determine prediction intervals for kinematic state.

\begin{table}[hb]
\begin{center}
\caption{Machine Learning Approaches}\label{tb:ML}
\begin{tabular}{ccccc}
Type & Method & P.P. & Inte. &Linear\\\hline
\multirow{2}*{Linear} & LR & Low & High & \cmark\\ & LASSO & Low & High & \cmark\\ \cline{2-5} \multirow{3} * {Ensemble} & RF & High & Low & \xmark\\  & GB & High & Low & \xmark\\  & XGB & High & Low & \xmark\\ \cline{2-5} \multirow{2} * {Kernel} & SVR & Medium & Medium & \xmark\\  & GP & High & Medium & \xmark\\ \cline{2-5} NN & MLP & High & Low &\xmark\\
 \hline
\end{tabular}
\end{center}
\begin{tablenotes}
  \small
  \item P.P.: predictive power, Inte.: interpretability, LR: linear regression, RF: random forest, GB: gradient boosting, XGB: XGBoost, SVR: support vector regressor, GP: Gaussian process, NN: neural network, MLP: multi-layer perceptron.
\end{tablenotes}
\end{table}

CP allows us to produce prediction sets with a predefined confidence level, ensuring that true values lie within these sets with a specified probability. The key of CP is to compute the \textit{nonconformity score}, which measures how different a new sample $(\mathbf{u}_{new}, \mathbf{x}_{new})$ is compared to a set of previously observed samples (typically the training set). Therefore, to establish a CP model, a nonconformity measure $\delta$ (e.g., absolute error between groundtruth and prediction) needs to be defined to assign a nonconformity score to each sample in  dataset $\mathcal{D}$. Following this, CP methods resort to $p$-values calculated by nonconformity scores to produce prediction sets. Specifically, the $p$-value of a new sample $(\mathbf{u}_{new}, \mathbf{x}_{new})$ can be computed by:
\begin{equation}\label{eq_4}
    p(\mathbf{x}_{new}) = \frac{|\{(\mathbf{u}_i,\mathbf{x}_i)\in\mathcal{D}:\delta(\mathbf{u}_i,\mathbf{x}_i)\geq \delta(\mathbf{u}_{new}, \mathbf{x}_{new})\}|+1}{N+1},
\end{equation}
Eq.~\ref{eq_4} implies that the prediction set for the new sample $\mathbf{u}_{new}$ includes all candidate labels for which the $p$-value exceeds a predefined significance level. In this context, we also notice that though CP and quantile regression (QR)~\citep{romano2019conformalized} share some common similarities on generating prediction intervals, the key difference is that CP provides a statistically guaranteed coverage probability for those intervals, regardless of the underlying data distribution. GP and Bayesian Neural Network (BNN), on the other hand, are probabilistic models that offer uncertainty estimates based on their inherent structure and prior assumptions. CP is a post-processing technique, meaning it can be applied to the output of any existing model, while GP and BNN are integrated into the model building process.

Building on CP, Split Conformal Prediction (SCP) improves the computational efficiency by avoiding repeated calculations of the nonconformity score for the given dataset. Instead, it is split into a training set and a calibration set. Analogously, the training set is used to fit a predictive model, while the calibration set is adopted for producing nonconformity scores, which measure the discrepancy between the groundtruth and predictions. Denote by $\mathcal{D}_{train}$ and $\mathcal{D}_{cal}$ the training and calibration sets such that $\mathcal{K}_f$ is trained on $\mathcal{D}_{train}$. Subsequently, its performance is evaluated on $\mathcal{D}_{cal}$ by calculating the nonconformity scores for each sample $(\mathbf{u}, \mathbf{x})\in\mathcal{D}_{cal}$. The nonconformity measure $\delta$ for any sample is defined as:
\begin{equation}\label{eq_5}
    \delta(\mathbf{u}, \mathbf{x}) = |\mathbf{x}-\mathcal{K}_f(\mathbf{u};\theta)| = |\mathbf{x}-\hat{\mathbf{x}}|.
\end{equation}
The score indicates the absolute error between the true value $\mathbf{x}$ and the predicted value $\hat{\mathbf{x}}$, which implies that the larger $\delta$, the higher nonconformity between the observed outcome and the model's prediction. 

Once the nonconformity scores are calculated for samples in $\mathcal{D}_{cal}$, we can select a quantile of these scores to establish the prediction intervals for new samples. Let $\mu$ be the quantile which corresponds to the desired confidence level $1-\alpha$, where $\alpha\in(0,1)$ is a predefined percentage value. For example, if $\alpha$ is set as 0.1, then at least 90\% of samples in $\mathcal{D}_{cal}$ will have nonconformity scores less than or equal to $\mu$, securing a guarantee that prediction intervals will cover true values with confidence level 90\%. The prediction interval $\mathcal{R}_{1-\alpha}$ for a new command $\mathbf{u}_{new}$ can be written as:
\begin{equation}\label{eq_6}
    \mathcal{R}_{1-\alpha} (\mathbf{u}_{new}) = [\mathcal{K}_f(\mathbf{u}_{new};\theta)-\mu,\mathcal{K}_f(\mathbf{u}_{new};\theta)+\mu] 
\end{equation}
The interval in Eq.~\ref{eq_6} indicates the range of possible values for the position variable $\mathbf{x}_{new}$ that complies with the nonconformity scores calculated from $\mathcal{D}_{cal}$. Note that SCP assumes the exchangeability among various sample, which is a severe and often inapplicable limitation in some cases such as timeseries data. To relax the exchangeability assumption in such applications, we can leverage the Weighted CP (WCP) developed recently in \citep{barber2023conformal} However, given our experimental data, the actuation-to-position mapping is time-independent, so we use classical SCP.

\begin{figure}
    \centering
    \includegraphics[width=0.9\linewidth]{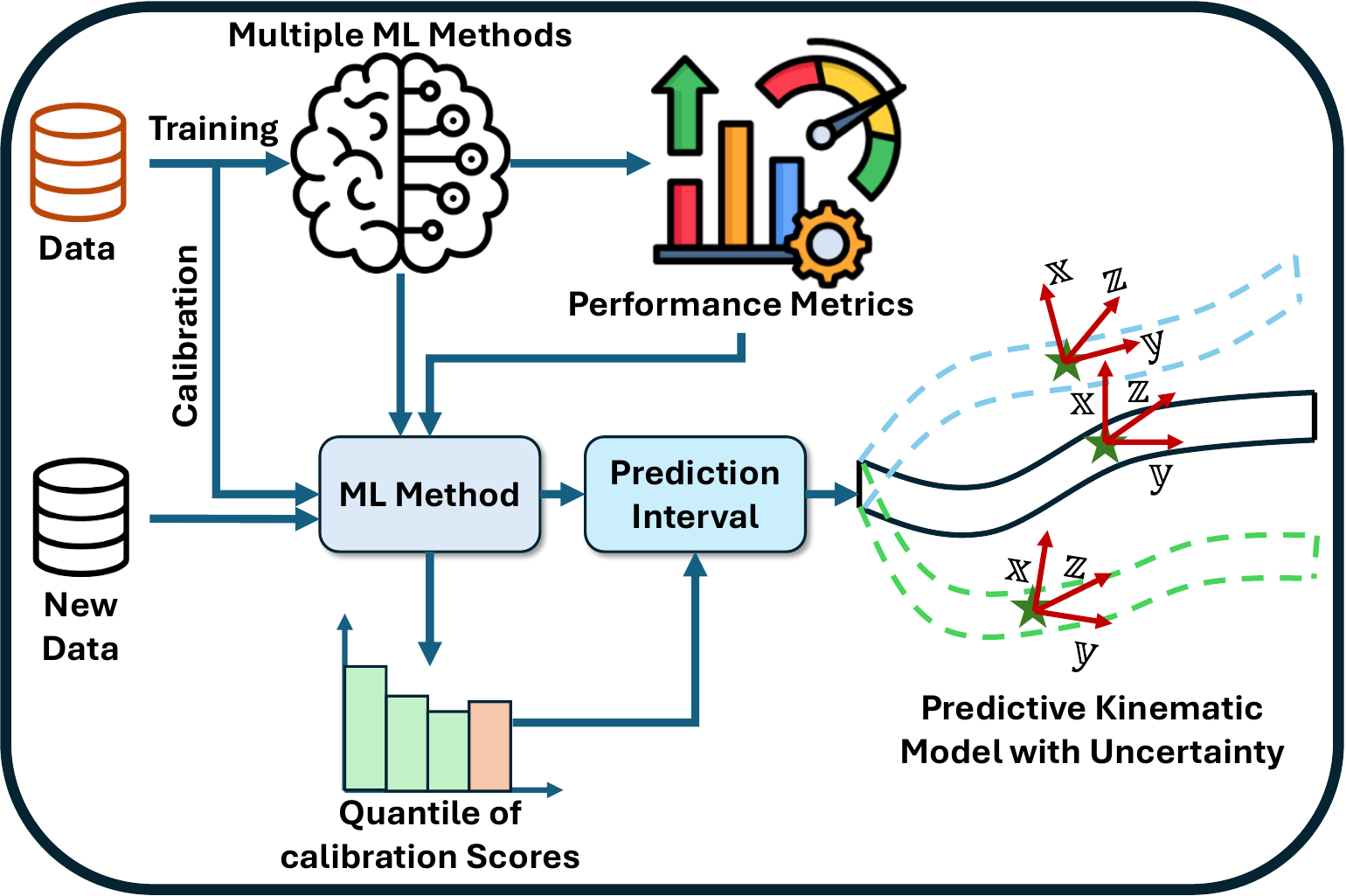}
    \caption{Schematic diagram of conformal kinematic modeling for soft robots: predictions with uncertainty.}
    \label{fig:ckm_sr}
\end{figure}

As shown in Figure~\ref{fig:ckm_sr}, given the data $\mathcal{D}$, we split it into the training set $\mathcal{D}_{train}$ and the calibration set $\mathcal{D}_{cal}$ (in practice, this can be the validation dataset). A pool of ML model candidates are trained on $\mathcal{D}_{train}$ based on the framework in Figure~\ref{fig:si_sr} to get performance metric values, which act as the criterion to select the most performant one for the subsequent conformal kinematic modeling. $\mathcal{D}_{cal}$ is then used to calibrate the quantile $\mu$ by computing nonconformity scores of all samples in it, with the predefined confidence level $\alpha$. Once a new sample with actuation commands is provided, the corresponding predictive position interval is obtained. 


\section{Experimental Results and Discussion}
\vspace{-0.1in}
\textbf{Dataset:} In this work, we resort to three datasets for validating the developed conformal kinematic modeling framework, including a pneumatic soft robot simulated by ABAQUS, a tendon-driven soft finger simulated by Elastica, and another tendon-driven soft finger in a real-world environment. These three datasets are from \textit{Yoon et al.}~\citep{5h7v-aq35-23}, and we refer interested readers to the associated research paper~\citep{yoon2024kinematics} for more details on how the datasets were generated and collected.

\begin{figure}
    \centering
    \includegraphics[width=1\linewidth]{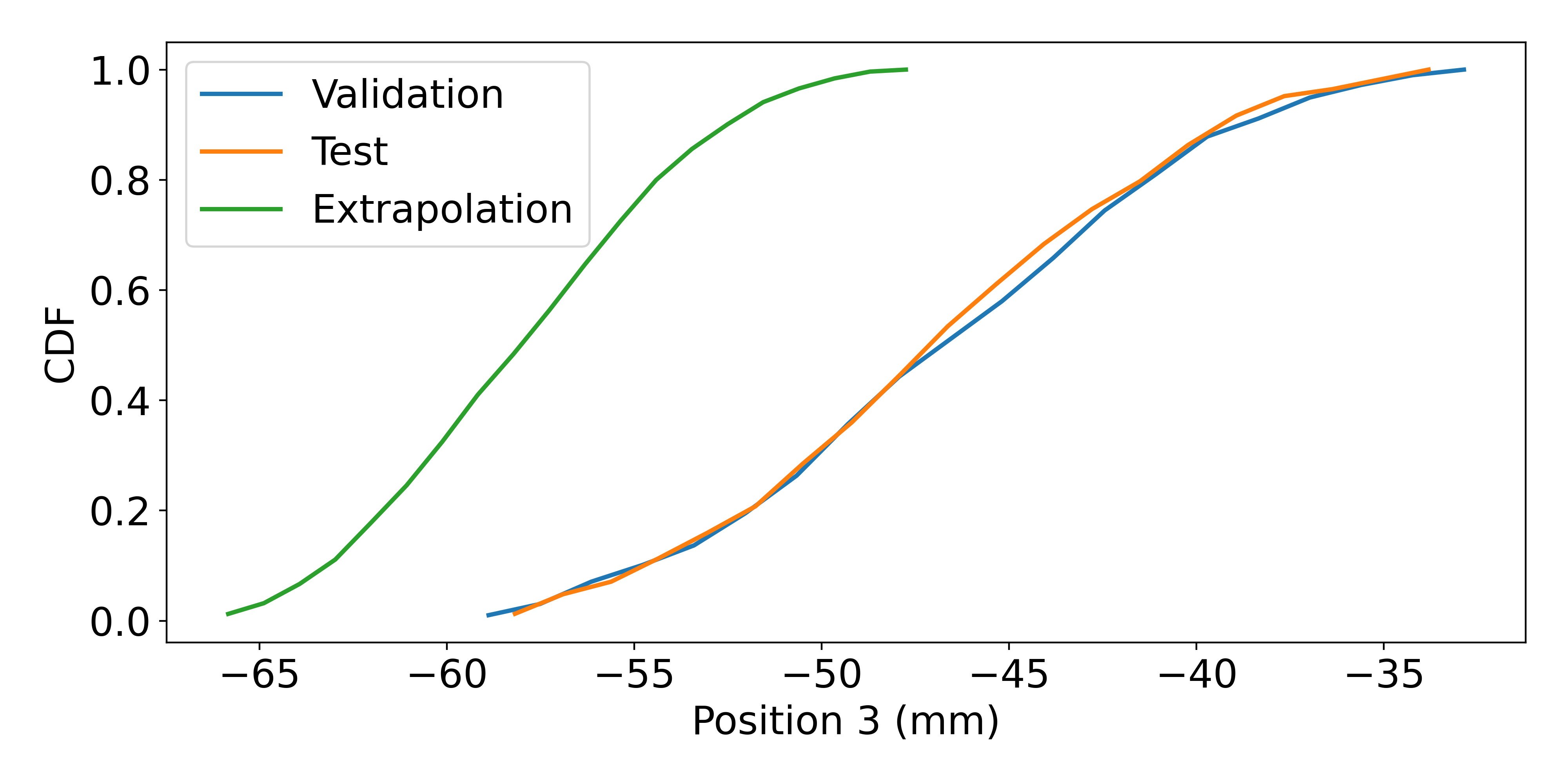}
    \caption{Cumulative Distribution Function for ABAQUS.}
    \label{fig:cdf_curve_aba}
\end{figure}

Each dataset consists of training, validation, testing and extrapolation data subsets (Table~\ref{tb:data}). The difference between these data subsets lies in data distributions. Typically, data distributions between training and validation or testing are much more similar than that between training and extrapolation, which is evidently validated in Figure~\ref{fig:cdf_curve_aba} (distribution from the other two datasets exhibit a similar trend). The plot shows the cumulative distribution function for output Position 3 (a selected coordinate) across validation, testing, and extrapolation data. Significant distribution drift between validation/testing and extrapolation explains inference difficulty on extrapolation data.

For each dataset, the input comprises motor commands, while the outputs are three-dimensional position coordinates. Each feature represents either one command or coordinate respectively. For example, the number of input features for ABAQUS is three, indicating three motor commands. Likewise, the number of output features indicate totally seven end effectors, as each end effector has three-dimensional position coordinates.
Therefore, the problem in Eq.~\ref{eq_3} becomes a multi-input and multi-output (MIMO) regression problem.

\begin{table}[hb]
\begin{center}
\caption{Data Description}\label{tb:data}
\begin{tabular}{ccccc}
Data & Subset & Size & \# of IF &\# of OF\\\hline
\multirow{4}*{ABAQUS (S)} & Tra. & 3160 & 3 & 21\\ & Val. & 395 & 3 & 21\\ & Tes. & 395 & 3 & 21 \\ &Ext. & 1694 & 3 & 21\\ \cline{2-5} \multirow{4}*{ELASTICA (S)} & Tra. & 1120 & 2 & 3\\ & Val. & 140 & 2 & 3\\ & Tes. & 140 & 2 & 3 \\ &Ext. & 600 & 2 & 3\\ \cline{2-5} \multirow{4} * {FINGER (R)} & Tra. & 704 & 2 & 3\\  & Val. & 88 & 2 & 3\\  & Tes. & 89 & 2 & 3 \\& Ext. & 378 & 2 & 3\\
 \hline
\end{tabular}
\end{center}
\begin{tablenotes}
  \small
  \item S: simulation, R: real-world, Tra.: training, Val.: validation, Tes.: test, Ext: extrapolation, IF: input features, OF: output features.
\end{tablenotes}
\end{table}

\textbf{Metrics:} To assess machine learning models' performance, we use the metrics of root mean squared error (RMSE: $\sqrt{\frac{1}{n}\sum_{i=1}^n(\mathbf{x}_i-\hat{\mathbf{x}}_i)^2}$), mean absolute error (MAE: $\frac{1}{n}\sum_{i=1}^n|\mathbf{x}_i-\hat{\mathbf{x}}_i|$), $R^2$ ($1-\frac{\sum_{i=1}^n(\mathbf{x}_i-\hat{\mathbf{x}}_i)^2}{\sum_{i=1}^n(\mathbf{x}_i-\bar{\mathbf{x}}_i)^2}$), and mean absolute percentage error (MAPE: $\frac{1}{n}\sum_{i=1}^n|\frac{\mathbf{x}_i-\hat{\mathbf{x}}_i}{\mathbf{x}_i}|*100\%$).
$\mathbf{x}_i$ is the true value, $\hat{\mathbf{x}}_i$ is the prediction, and $\bar{\mathbf{x}}_i$ is the average of true values. 
To evaluate conformal kinematic modeling performance, we utilize two metrics, including coverage and Mean Winkler Interval Score~\citep{winkler1972decision}. The coverage is the empirical percentage of true values contained in the prediction sets for either testing or extrapolation data. For each true scalar value $y$, Winkler Interval Score is defined as follows:
\begin{equation}
W_\alpha=
    \begin{cases}
        v_u-v_l+\frac{2}{\alpha}(v_l-y) & \text{if $y<v_l$}\\
        v_u-v_l & \text{$v_l\leq y\leq v_u$}\\
        v_u-v_l+\frac{2}{\alpha}(y-v_u) & \text{if $y>v_u$}
    \end{cases},
\end{equation}
where $v_u$ is the upper prediction interval, $v_l$ is the lower prediction interval, and $\alpha$ is the the probability of a prediction interval failing to contain the true value.
Given a sequence of each position feature, the Mean Winkler Interval Score is the mean of Winkler Interval Scores of all true values. In this work, we set $\alpha=0.1$ and apply this to the quantile setting for all uncertainty quantification models. Hyperparameters were manually tuned for near-optimal performance, though optimization methods could yield further improvements. $\uparrow$ in this work means that the higher value is the better, while $\downarrow$ the lower the better. There are two sides for a targeted position such that a negative value in the following results means the opposite side relative to the original one.

\begin{figure}
\begin{subfigure}{1\columnwidth}
    \centering
    \includegraphics[width=\linewidth]{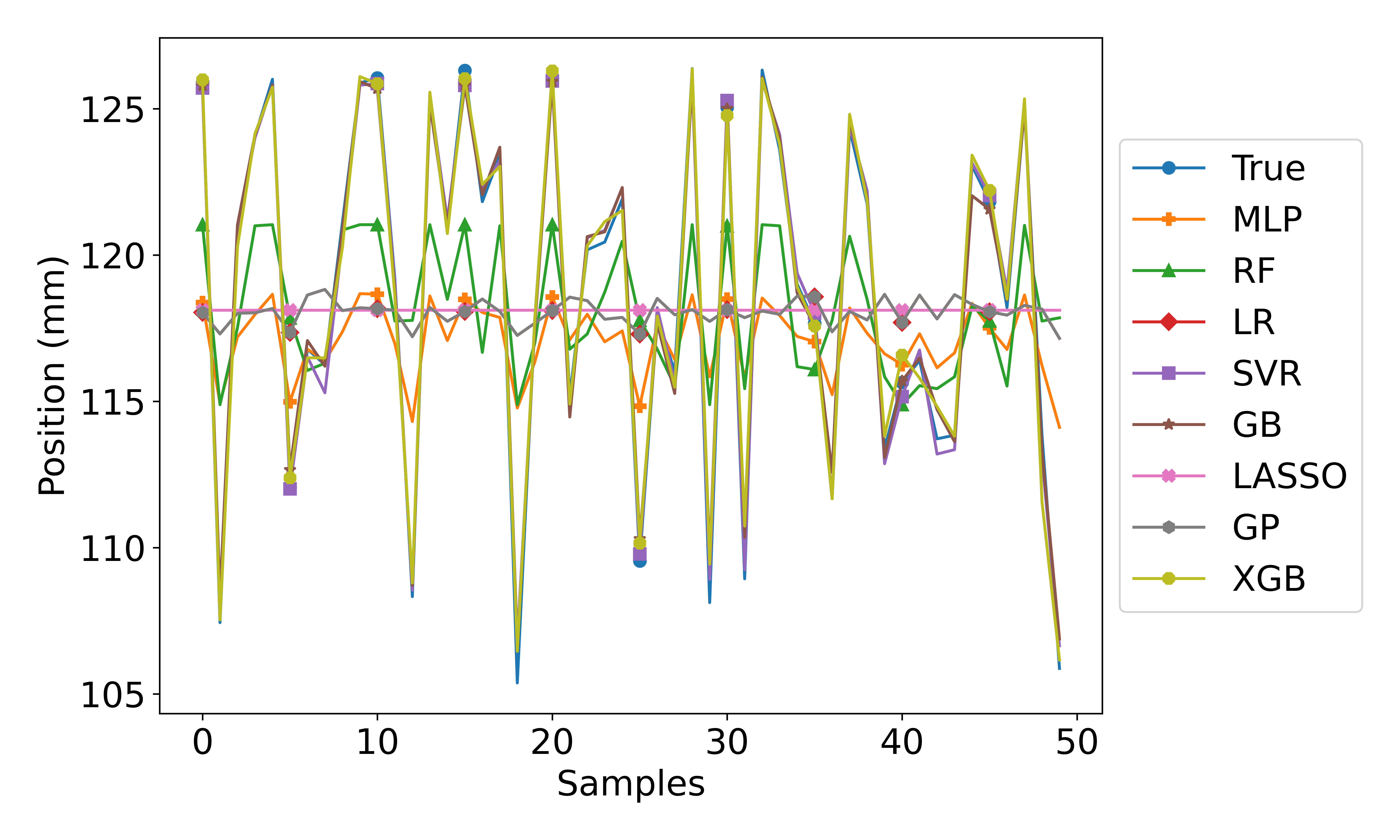}
    \caption{Testing data.}
\end{subfigure}
\begin{subfigure}{1\columnwidth}
    \centering
\includegraphics[width=\linewidth]{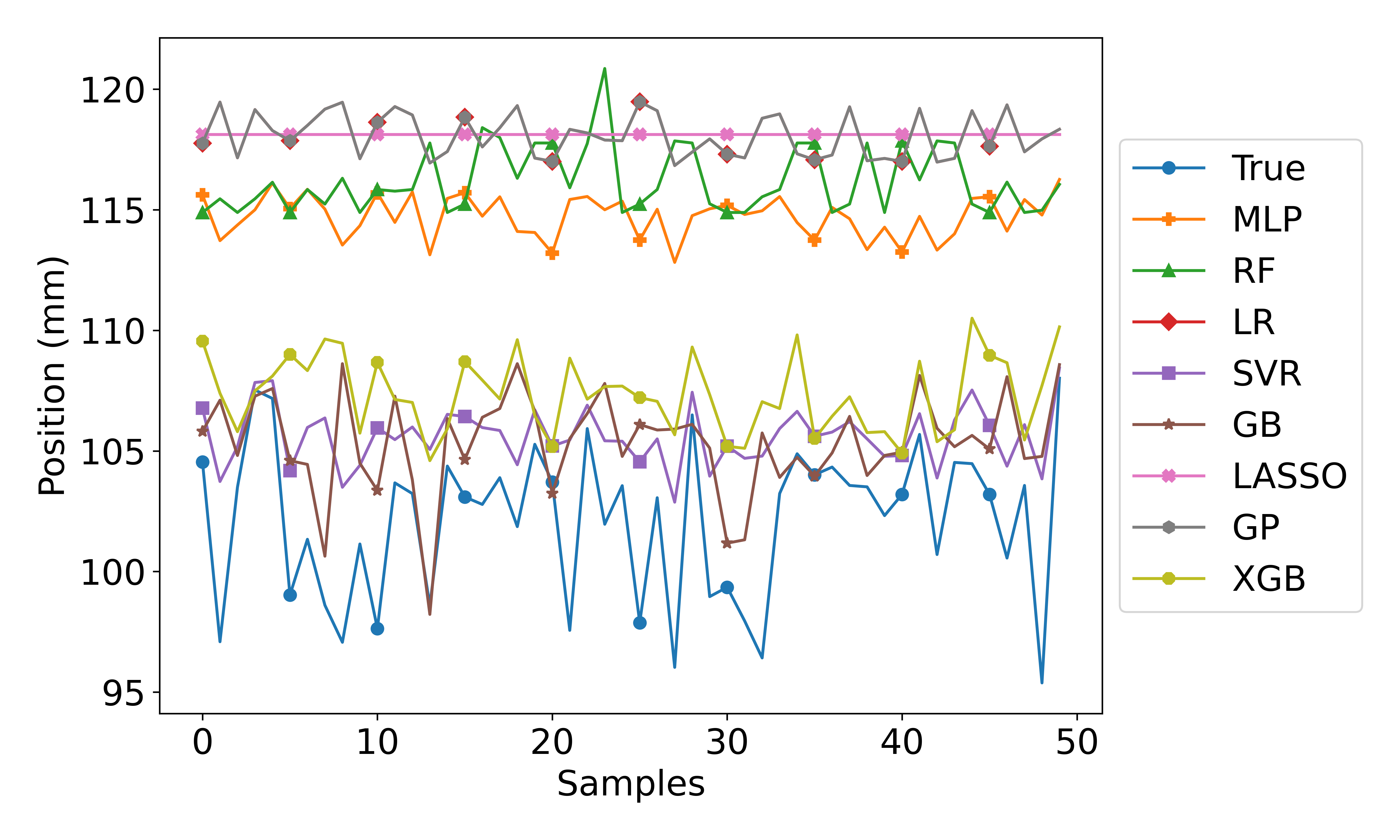}
    \caption{Extrapolation data.}
    
\end{subfigure}
    \caption{Position Curve between ground truth and predictions by different methods for Position 2 in the output for FINGER}\label{fig:fin_position_curve_test_data}
\end{figure}

\textbf{System Identification Performance.}
We evaluate machine learning models in Table~\ref{tb:ML} on the three datasets and show the summary of results in Tables~\ref{tb:sid_aba_test_performance} to~\ref{tb:sid_fin_test_performance}. Overall, the GB and XGB are the most robust performers generalizing trained models to testing and extrapolation phases among various models across different datasets. Though LR and GP are notably competitive to GB/XGB for ABAQUS, as shown in Figure~\ref{fig:fin_position_curve_test_data}, they perform poorly for the real-world data FINGER, particularly on the extrapolation data.
Despite the performance degradation for all models when testing on extrapolation data, GB/XGB remains more robust against distribution drifts in general. 

Remarkably, GB/XGB outshine other models when looking into RMSE, MAE, and MAPE for the testing data of ELASTICA (Table~\ref{tb:sid_ela_test_performance}). This is attributed to combining multiple weak learners to create a strong model and the regularization techniques to prevent overfitting and improve model generalization.
For different model categories, linear models rank the worst based on  results, which is also empirically supported by Figure~\ref{fig:fin_position_curve_test_data} (larger deviation from ground truth). In particular, for the FINGER data, LR and LASSO fail to learn the underlying relationship between motor commands and positions. LASSO performs the worst because its penalty forces some of coefficients quickly to zero, leading to the poor sparsity that causes inability to learn the complex relationship.

\begin{figure}
\begin{subfigure}{1\columnwidth}
    \centering
    \includegraphics[width=\linewidth]{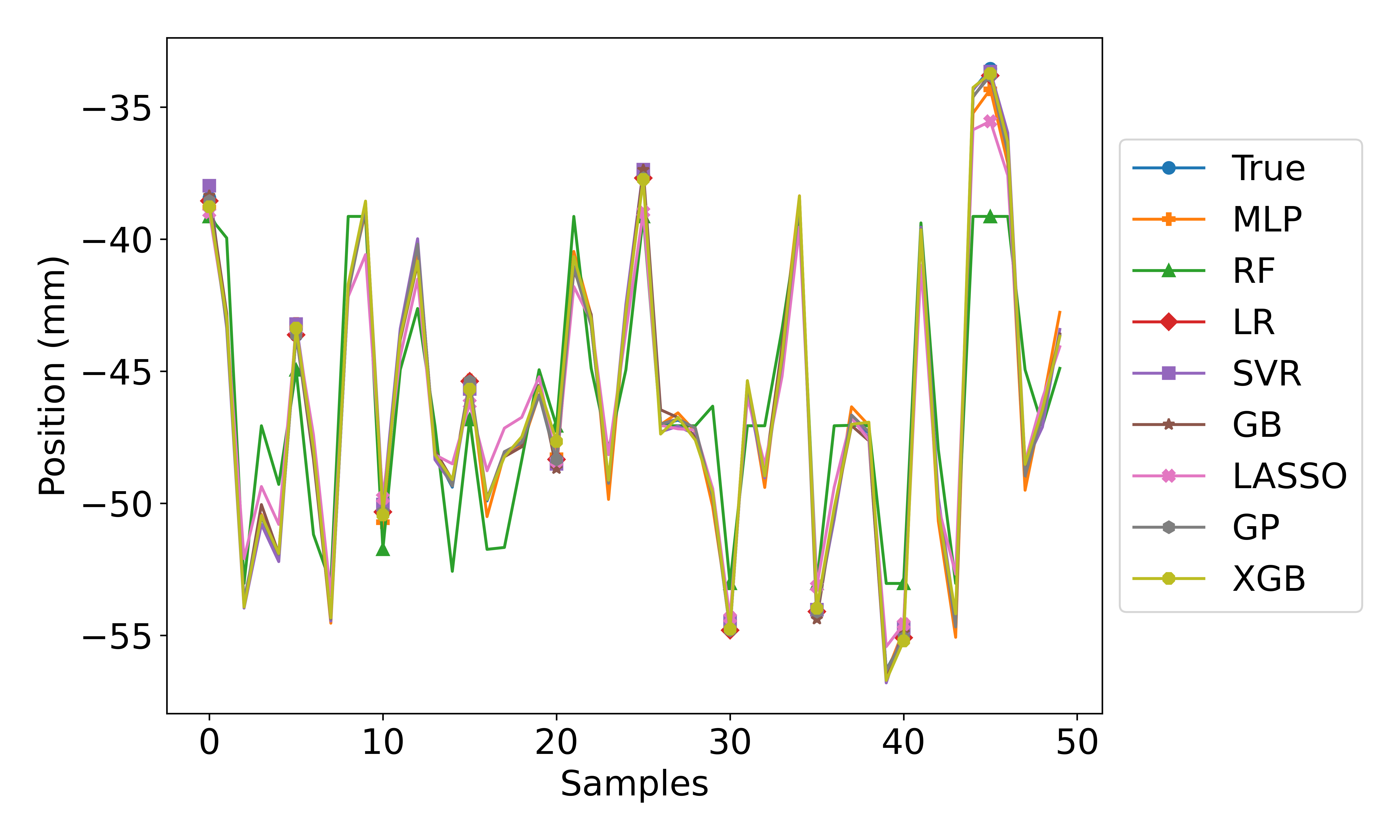}
    \caption{Testing data.}
\end{subfigure}
\begin{subfigure}{1\columnwidth}
    \centering
\includegraphics[width=\linewidth]{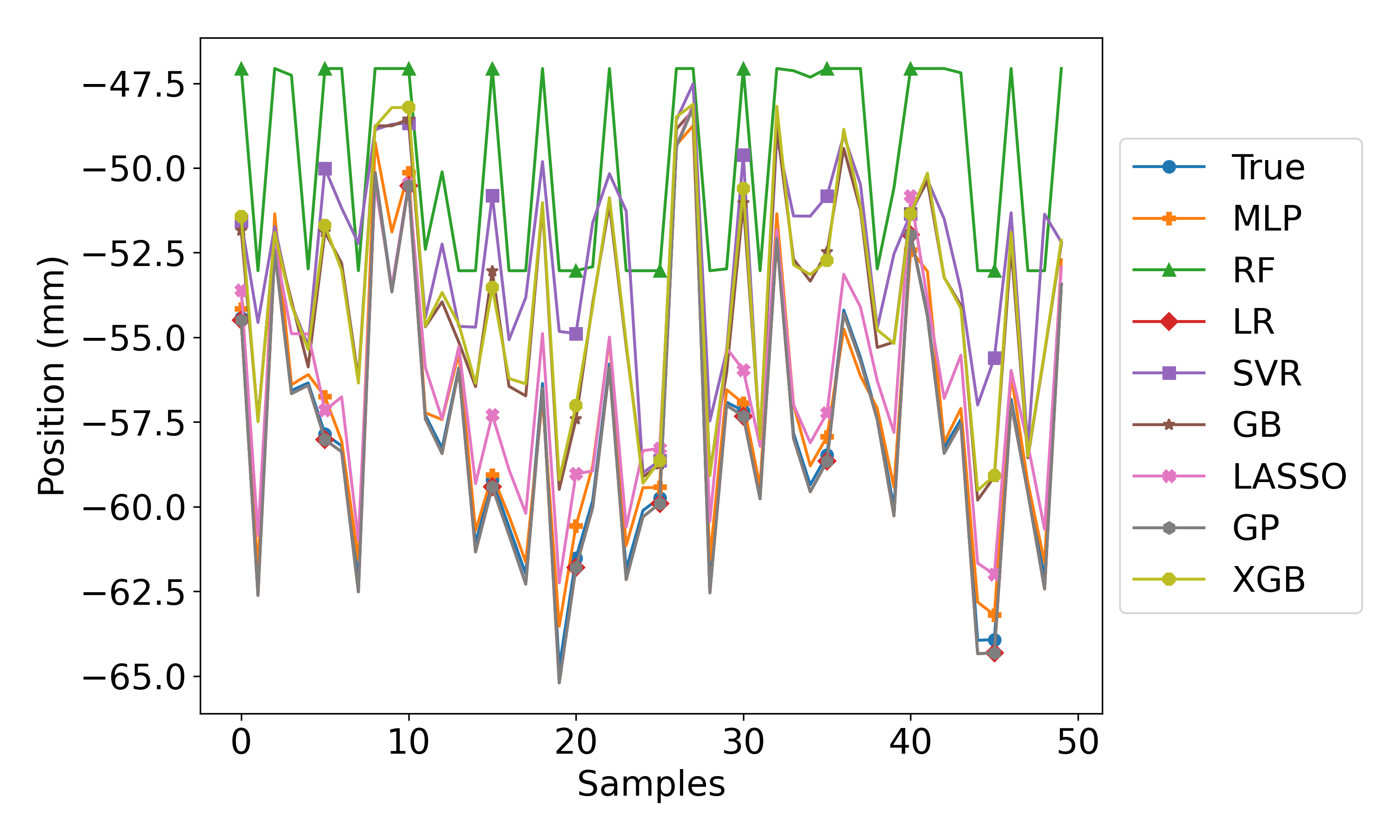}
    \caption{Extrapolation data.}
\end{subfigure}
\caption{Position Curve between ground truth and predictions by different methods for Position 3 in the output for ABAQUS.}
\label{fig:aba_position_curve_test_data}
\end{figure}

Though RF is also an ensemble method as GB and XGB, its performance is much worse across diverse data due to its lack of regularization techniques, such as in Figures~\ref{fig:aba_position_curve_test_data} and~\ref{fig:fin_position_curve_test_data}. Also, RF combines results from multiple trees that most likely saturate in each prediction. Kernel-based methods such as SVR and GP degrade their performance when moving from testing to extrapolation data owing to the data distribution drift. When turning to MLP, we have observed that it performs moderately across all methods. Compared to other models, MLP has more number of parameters, thus requiring a large amount of data for a well-trained model. 
To summarize, ensemble methods like GB and XGB are the best model candidates for kinematic modeling in soft robots. Thus, we select GB as the model to proceed on the conformal kinematic modeling.
\begin{table}[hb]
\begin{center}
\caption{Comparison on ABAQUS}\label{tb:sid_aba_test_performance}
\begin{tabular}{ccccc}
\multicolumn{5}{c}{Testing} \\\hline
Method & RMSE ($\downarrow$) & MAE ($\downarrow$) & $R^2$ ($\uparrow$) &MAPE ($\downarrow$)\\\hline
LR & 2.89 &\textbf{1.10}  &0.991 &\textbf{0.32\%}\\ LASSO & 6.09 & 4.80 & 0.966& 2.27\%\\  RF & 14.97 & 12.27 & 0.810 & 5.76\%\\   GB & \textbf{2.05} &  1.37& 0.996&0.64\%\\  XGB & 2.08 & 1.41 &\textbf{0.997} &0.63\%\\  SVR & 2.71 & 1.45 &0.994 &0.65\%\\   GP & 2.89 & \textbf{1.10} & 0.991&\textbf{0.32\%}\\ MLP & 3.13 & 1.90  & 0.899& 0.94\%\\
 \hline
 \multicolumn{5}{c}{Extrapolation}\\\hline
 Method & RMSE ($\downarrow$) & MAE ($\downarrow$) & $R^2$ ($\uparrow$) &MAPE ($\downarrow$)\\\hline
LR & \textbf{3.29} &\textbf{2.02}  &0.971 &\textbf{0.32\%}\\ LASSO & 10.68 & 9.05 & 0.843& 2.53\%\\  RF & 33.98 & 30.10 & -0.824 & 10.08\%\\   GB & 17.73 &  15.73& 0.501&5.25\%\\  XGB & 17.38 & 15.33 &0.505 &5.17\%\\  SVR & 22.20 & 16.98 &0.178 &5.65\%\\   GP & \textbf{3.29} & 2.03 & \textbf{0.972}&\textbf{0.32\%}\\ MLP & 4.87 & 3.80  & 0.866& 1.14\%\\
 \hline
\end{tabular}
\end{center}
\end{table}


\begin{table}[hb]
\begin{center}
\caption{Comparison on ELASTICA}\label{tb:sid_ela_test_performance}
\begin{tabular}{ccccc}
\multicolumn{5}{c}{Testing} \\\hline
Method & RMSE ($\downarrow$) & MAE ($\downarrow$) & $R^2$ ($\uparrow$) &MAPE ($\downarrow$)\\\hline
LR & 0.86 &0.77  &0.693 &1.39\%\\ LASSO & 2.05 &  1.82& 0.673& 8.17\%\\  RF & 6.41 & 5.03 & 0.649 & 90.02\%\\   GB & 0.007 &  0.005& \textbf{0.999}&5.57e-3\%\\  XGB & \textbf{0.0005} & \textbf{0.0003} &\textbf{0.999} &\textbf{2.53e-3\%}\\  SVR & 1.00 & 0.88 &0.993 &9.69\%\\   GP & 0.86 & 0.77 & 0.693&1.39\%\\ MLP & 2.21 & 1.77  & 0.772& 9.67\%\\
 \hline
 \multicolumn{5}{c}{Extrapolation}\\\hline
 Method & RMSE ($\downarrow$) & MAE ($\downarrow$) & $R^2$ ($\uparrow$) &MAPE ($\downarrow$)\\\hline
LR & 3.59 &\textbf{3.30}  &-2.038 &4.35\%\\ LASSO & 4.50 &  4.18& -1.965& 7.30\%\\  RF & 16.29 & 14.92 & -2.303 & 63.41\%\\   GB & 4.00 &  \textbf{3.30}& \textbf{-0.048}&10.33\%\\  XGB & 4.80 & 4.23 &-0.438&13.04\%\\  SVR &13.03 & 11.29 &-0.130 &33.54\%\\   GP & \textbf{3.58} & \textbf{3.30} & -2.038&\textbf{4.32\%}\\ MLP & 6.36 & 5.78  & -1.346& 15.28\%\\
 \hline
\end{tabular}
\end{center}
\end{table}


\begin{table}[hb]
\begin{center}
\caption{Comparison FINGER}\label{tb:sid_fin_test_performance}
\begin{tabular}{ccccc}
\multicolumn{5}{c}{Testing} \\\hline
Method & RMSE ($\downarrow$) & MAE ($\downarrow$) & $R^2$ ($\uparrow$) &MAPE ($\downarrow$)\\\hline
LR & 4.33 &3.64 &0.659 &86.92\%\\ LASSO & 4.89 &  4.15& 0.629& 112.58\%\\  RF & 7.65 & 5.85 & 0.633 & 70.96\%\\   GB &1.59 &  1.25& 0.986&50.47\%\\  XGB & 1.80 & 1.41&0.983&\textbf{28.92\%}\\  SVR &\textbf{1.40}& \textbf{1.14} &\textbf{0.989} &42.80\%\\   GP & 4.32 & 3.64 & 0.659&86.92\%\\ MLP & 6.54 & 5.39  & 0.678& 85.82\%\\
 \hline
 \multicolumn{5}{c}{Extrapolation}\\\hline
 Method & RMSE ($\downarrow$) & MAE ($\downarrow$) & $R^2$ ($\uparrow$) &MAPE ($\downarrow$)\\\hline
LR & 10.10 &9.30  &-6.678 &114.73\%\\ LASSO & 9.39 &  8.44& -6.647& 150.43\%\\  RF & 19.18 & 16.47 & -5.614 & 124.62\%\\   GB &\textbf{5.40} &  \textbf{4.28}& \textbf{0.215}&71.80\%\\  XGB & 5.91 &4.83&-0.312&64.89\%\\  SVR &9.12& 7.13 &0.251 &70.28\%\\   GP & 10.09 & 9.29 & -6.678&114.64\%\\ MLP & 12.02 & 10.41  & -3.971& \textbf{39.54\%}\\
 \hline
\end{tabular}
\end{center}
\end{table}


\begin{figure}
\begin{subfigure}{1\columnwidth}
    \centering
    \includegraphics[width=\linewidth]{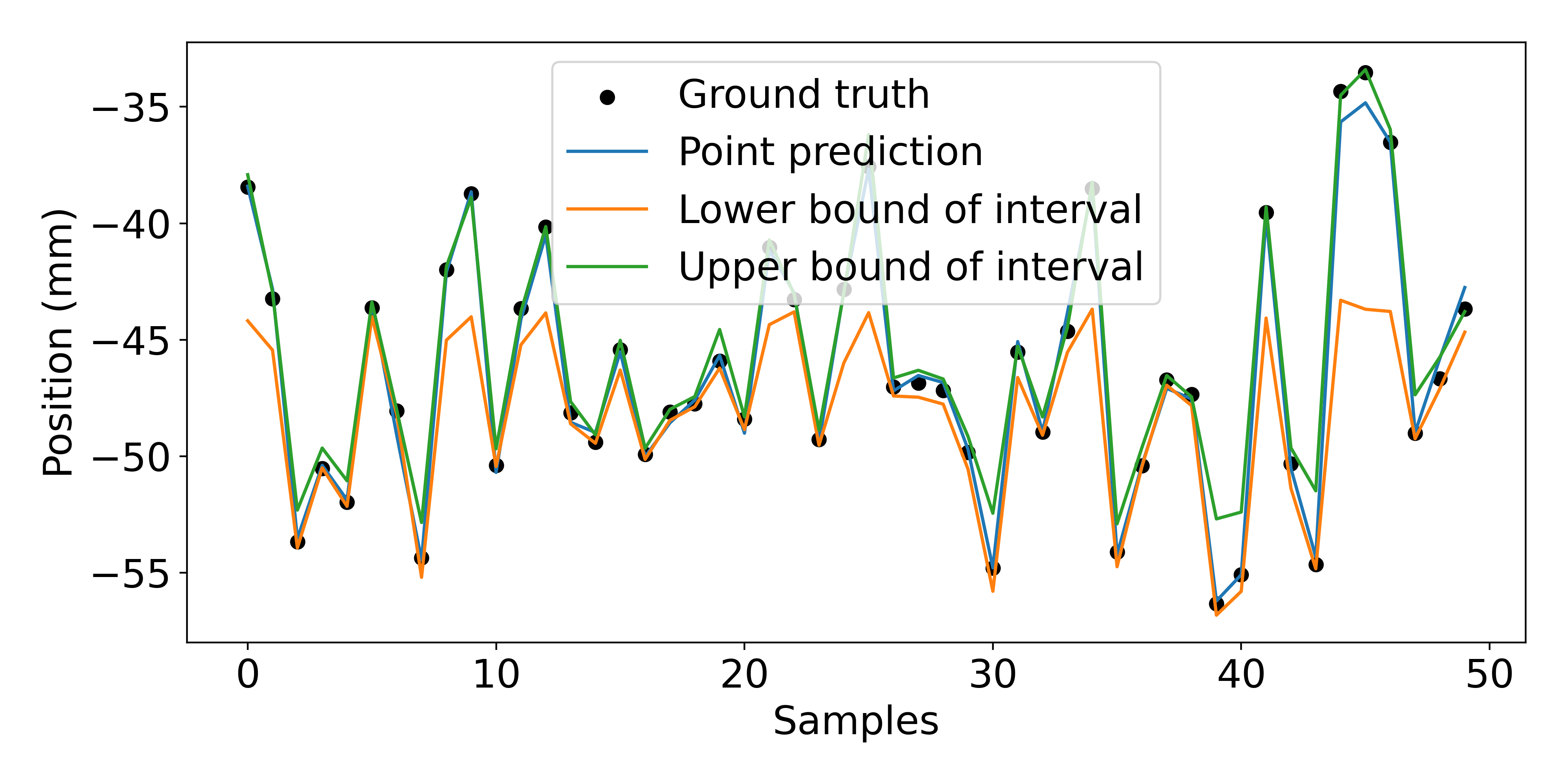}
    \caption{QR.}
\end{subfigure}
\begin{subfigure}{1\columnwidth}
        \centering
    \includegraphics[width=\linewidth]{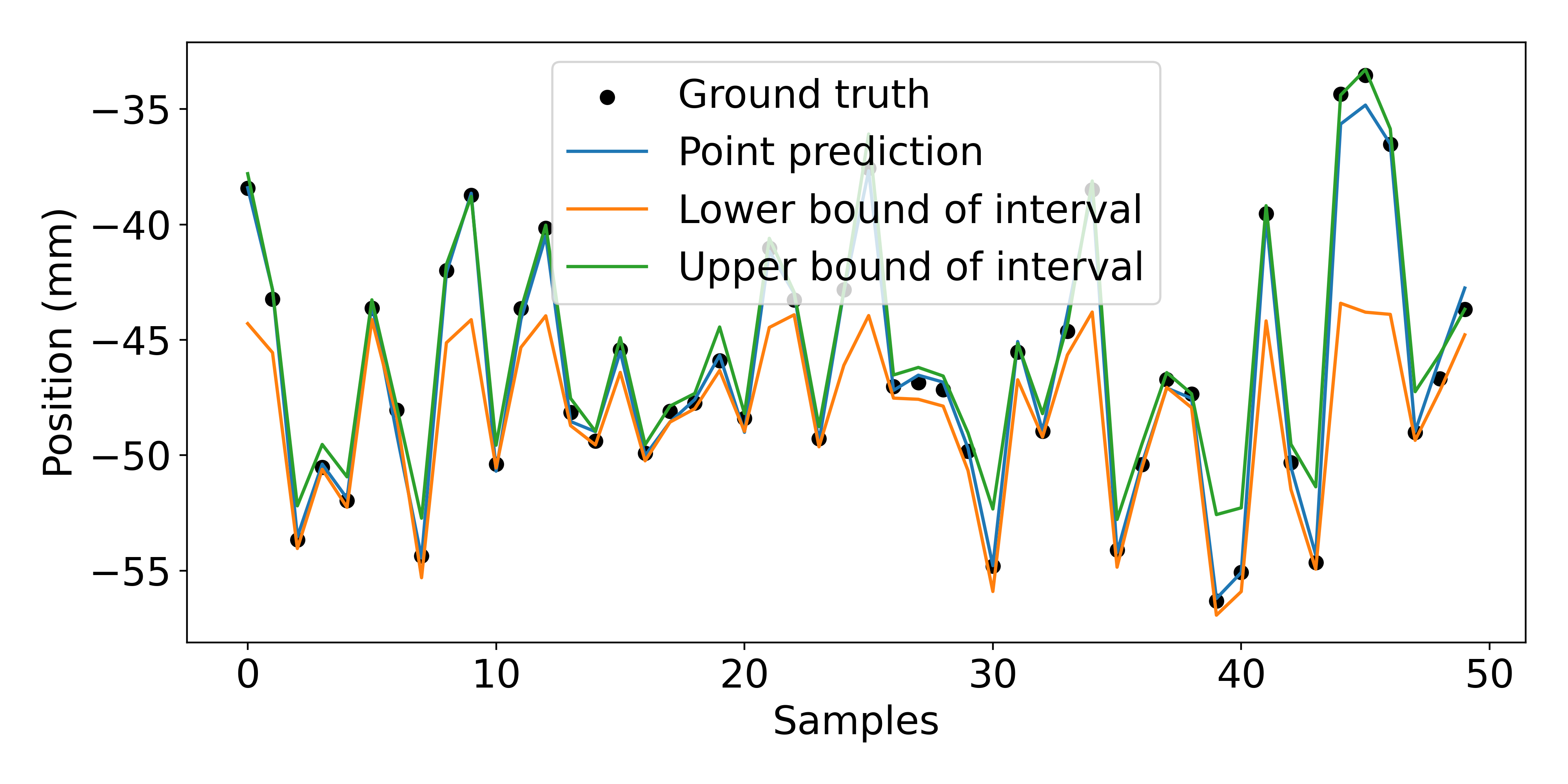}
    \caption{CQR.}
\end{subfigure}
\begin{subfigure}{1\columnwidth}
        \centering
    \includegraphics[width=\linewidth]{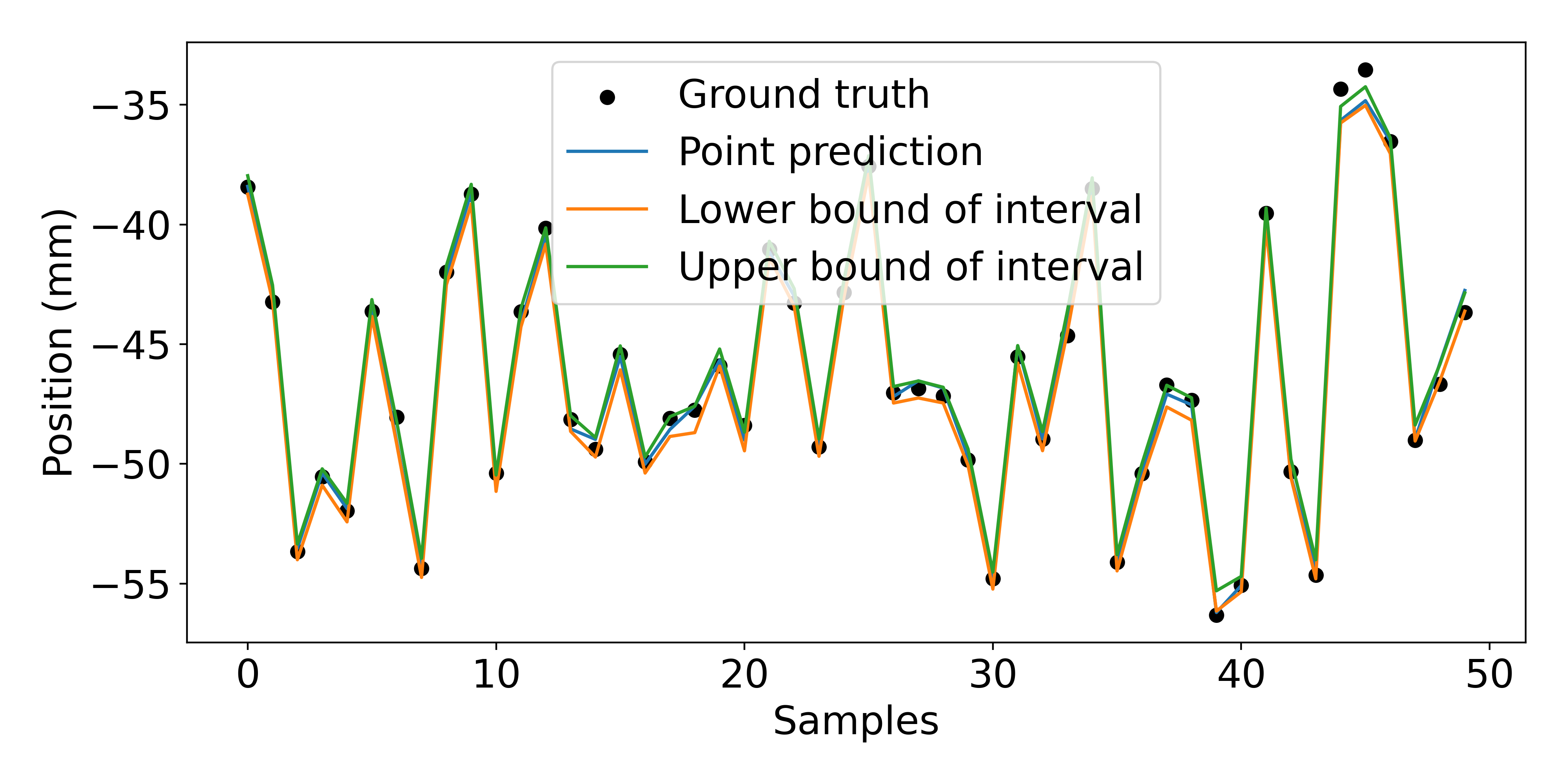}
    \caption{SCP (MAPIE).}
\end{subfigure}
\begin{subfigure}{1\columnwidth}
        \centering
    \includegraphics[width=\linewidth]{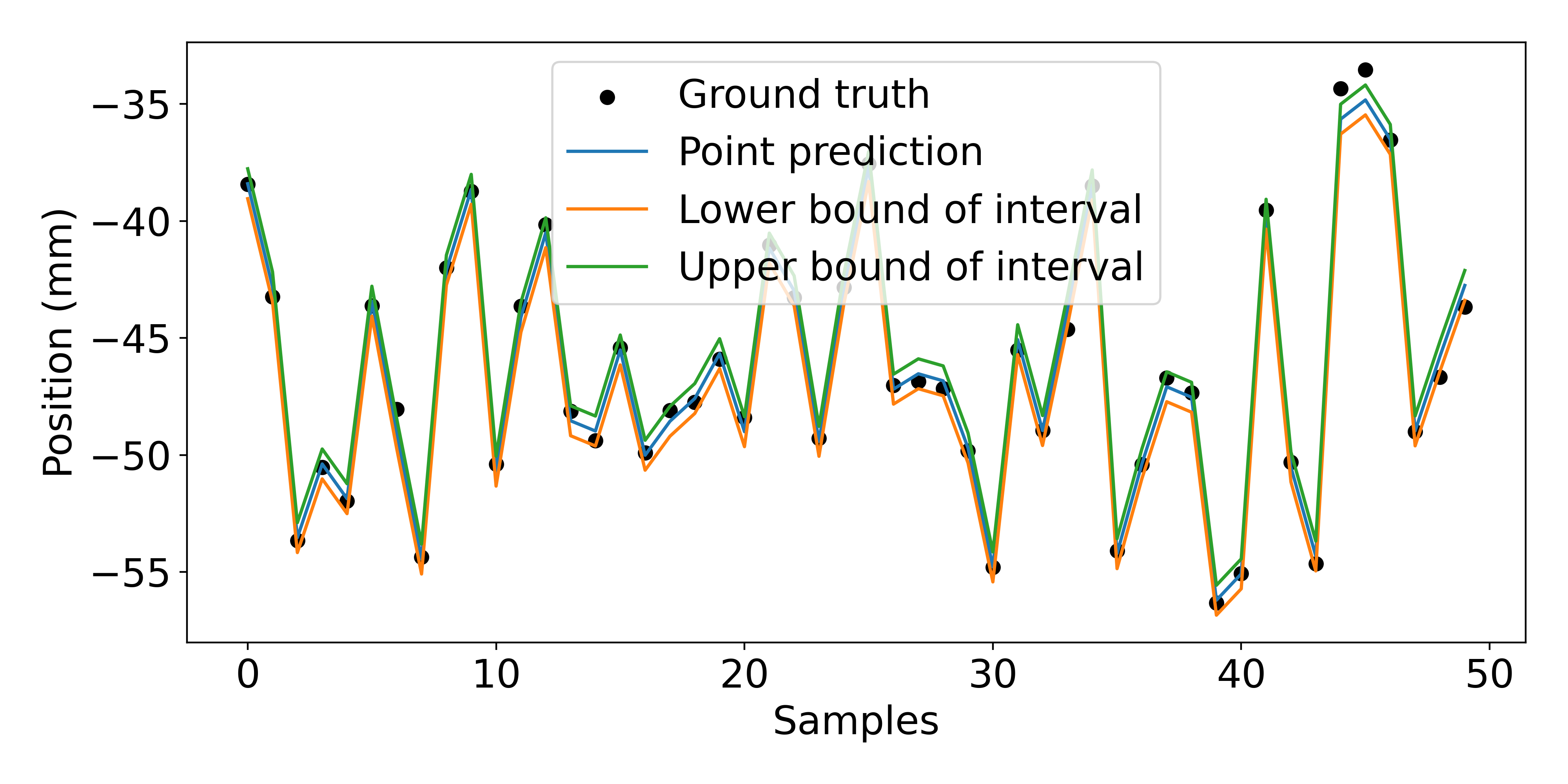}
    \caption{SCP (CREPES).}
\end{subfigure}
\caption{Conformal prediction performance with ABAQUS testing data for Position 3.}
    \label{fig:qr_test}
\end{figure}

\begin{table}[hb]
\begin{center}
\caption{Comparison between conformal prediction and quantiled regression models}\label{tb:conformal_datasets}
\begin{tabular}{cccccc}
Data&Metric & QR & CQR & SCP (M) &SCP (C)\\\hline
\multirow{2}*{ABA.} & Coverage ($\uparrow$) & 0.80 &0.93 &\textbf{0.98}&0.93\\ & Winkler score ($\downarrow$) & 3.12 & 3.03 & 1.96& \textbf{1.85}\\ \hline \multirow{2}*{ELA.} & Coverage ($\uparrow$) & 0.81 &0.91 &\textbf{1.00}&0.99\\ & Winkler score ($\downarrow$) & 4.83 & 4.83 & 2.18& \textbf{2.06}\\\hline \multirow{2}*{FIN.} & Coverage ($\uparrow$) & 0.78 &0.90 &\textbf{0.99}&0.83\\ & Winkler score ($\downarrow$)& 10.39 & 9.93 & 2.97& \textbf{2.84}\\
 \hline
\end{tabular}
\end{center}
\begin{tablenotes}
  \small
  \item ABA.: ABAQUS, ELA.: ELASTICA, FIN.: FINGER, M: MAPIE, C: CREPES.
\end{tablenotes}
\end{table}

\textbf{Conformal Prediction.}
Soft robots have structural compliance and viscoelasticity, hindering precise kinematic modeling. Therefore, quantifying kinematic modeling uncertainty becomes important.
As discussed before, SCP can provide us with a statistically valid prediction interval that informs us a certain probability that a true value is contained in that interval. To compare SCP to existing methods, we use QR and conformalized QR (CQR)~\citep{romano2019conformalized}. 
For SCP, we adopt two popular packages: MAPIE and CREPES~\citep{uddin2023applications}. Results in Table~\ref{tb:conformal_datasets} with testing data from all datasets show that SCP yields the best performance for coverage and Winkler score. The improvement (coverage/Winkler) for three different datasets (ABAQUS/ELASTICA/FINGER) is 23\%/41\%, 23\%/57\%, and 27\%/73\%. Compared to QR, SCP does not require any distributional assumptions to guarantee coverage. Additionally, QR intervals may not be calibrated, meaning the actual coverage might be off from the desired level. CQR outperforms QR, suggesting that calibration assists in prediction performance. When comparing two versions of SCP, we can observe that the one associated with MAPIE performs slightly better than that in CREPES, as the former is suitable for complex tasks where underlying relationships are highly nonlinear (Figure~\ref{fig:qr_test}). One immediate observation is that SCP produces narrower prediction intervals, empirically validating that conformal kinematic modeling for soft robots are more reliable than using the popular QR. 
\section{Conclusions and Future Directions}
\vspace{-0.1in}
For soft robot kinematics, ensemble methods (GB, XGB) showed robustness to data drift. Addressing unreliable modeling, we built conformal kinematic models via split conformal prediction. These models outperformed quantile regression, yielding higher coverage and lower mean Winkler scores. This framework enables reliable data-driven kinematics and efficient stochastic controller design.


\bibliography{ifacconf}

\begin{thebibliography}{17}
\providecommand{\natexlab}[1]{#1}
\providecommand{\url}[1]{\texttt{#1}}
\providecommand{\urlprefix}{URL }
\expandafter\ifx\csname urlstyle\endcsname\relax
  \providecommand{\doi}[1]{doi:\discretionary{}{}{}#1}\else
  \providecommand{\doi}{doi:\discretionary{}{}{}\begingroup \urlstyle{rm}\Url}\fi

\bibitem[{Abbasi et~al.(2020)Abbasi, Nekoui, Zareinejad, Abbasi, and Azhang}]{abbasi2020position}
Abbasi, P., Nekoui, M.A., Zareinejad, M., Abbasi, P., and Azhang, Z. (2020).
\newblock Position and force control of a soft pneumatic actuator.
\newblock \emph{Soft Robotics}, 7(5), 550--563.

\bibitem[{Arapi et~al.(2020)Arapi, Zhang, Averta, Catalano, Rus, Della~Santina, and Bianchi}]{arapi2020grasp}
Arapi, V., Zhang, Y., Averta, G., Catalano, M.G., Rus, D., Della~Santina, C., and Bianchi, M. (2020).
\newblock To grasp or not to grasp: an end-to-end deep-learning approach for predicting grasping failures in soft hands.
\newblock In \emph{2020 3rd IEEE International Conference on Soft Robotics (RoboSoft)}, 653--660. IEEE.

\bibitem[{Armanini et~al.(2023)Armanini, Boyer, Mathew, Duriez, and Renda}]{armanini2023soft}
Armanini, C., Boyer, F., Mathew, A.T., Duriez, C., and Renda, F. (2023).
\newblock Soft robots modeling: A structured overview.
\newblock \emph{IEEE Transactions on Robotics}, 39(3), 1728--1748.

\bibitem[{Barber et~al.(2023)Barber, Candes, Ramdas, and Tibshirani}]{barber2023conformal}
Barber, R.F., Candes, E.J., Ramdas, A., and Tibshirani, R.J. (2023).
\newblock Conformal prediction beyond exchangeability.
\newblock \emph{The Annals of Statistics}, 51(2), 816--845.

\bibitem[{Beaber et~al.(2024)Beaber, Liu, and Sun}]{beaber2024physics}
Beaber, S., Liu, Z., and Sun, Y. (2024).
\newblock Physics-guided deep learning enabled surrogate modeling for pneumatic soft robots.
\newblock \emph{IEEE Robotics and Automation Letters}.

\bibitem[{Dou et~al.(2021)Dou, Zhong, Cao, Shi, Peng, and Jiang}]{dou2021soft}
Dou, W., Zhong, G., Cao, J., Shi, Z., Peng, B., and Jiang, L. (2021).
\newblock Soft robotic manipulators: Designs, actuation, stiffness tuning, and sensing.
\newblock \emph{Advanced Materials Technologies}, 6(9), 2100018.

\bibitem[{Jitosho et~al.(2023)Jitosho, Lum, Okamura, and Liu}]{jitosho2023reinforcement}
Jitosho, R., Lum, T.G.W., Okamura, A., and Liu, K. (2023).
\newblock Reinforcement learning enables real-time planning and control of agile maneuvers for soft robot arms.
\newblock In \emph{Conference on Robot Learning}, 1131--1153. PMLR.

\bibitem[{Pinskier and Howard(2022)}]{pinskier2022bioinspiration}
Pinskier, J. and Howard, D. (2022).
\newblock From bioinspiration to computer generation: Developments in autonomous soft robot design.
\newblock \emph{Advanced Intelligent Systems}, 4(1), 2100086.

\bibitem[{Romano et~al.(2019)Romano, Patterson, and Candes}]{romano2019conformalized}
Romano, Y., Patterson, E., and Candes, E. (2019).
\newblock Conformalized quantile regression.
\newblock \emph{Advances in neural information processing systems}, 32.

\bibitem[{Seo et~al.(2024)Seo, Na, Kim, Kim, Kim, Chun, and Han}]{seo2024soft}
Seo, S., Na, H.M., Kim, J.Y., Kim, D., Kim, D., Chun, K.Y., and Han, C.S. (2024).
\newblock Soft and integrable multimodal artificial mechanoreceptors toward human sensor of skin.
\newblock \emph{Advanced Functional Materials}, 2414489.

\bibitem[{Uddin and Lofstrom(2023)}]{uddin2023applications}
Uddin, N. and Lofstrom, T. (2023).
\newblock Applications of conformal regression on real-world industrial use cases using crepes and mapie.
\newblock In \emph{Conformal and Probabilistic Prediction with Applications}, 147--165. PMLR.

\bibitem[{Wang et~al.(2021)Wang, Sun, Zhao, and Zuo}]{wang2021highly}
Wang, S., Sun, Z., Zhao, Y., and Zuo, L. (2021).
\newblock A highly stretchable hydrogel sensor for soft robot multi-modal perception.
\newblock \emph{Sensors and Actuators A: Physical}, 331, 113006.

\bibitem[{Wang et~al.(2024)Wang, Wang, Ge, Duan, Chen, and Wen}]{wang2024perceived}
Wang, Y., Wang, G., Ge, W., Duan, J., Chen, Z., and Wen, L. (2024).
\newblock Perceived safety assessment of interactive motions in human--soft robot interaction.
\newblock \emph{Biomimetics}, 9(1), 58.

\bibitem[{Winkler(1972)}]{winkler1972decision}
Winkler, R.L. (1972).
\newblock A decision-theoretic approach to interval estimation.
\newblock \emph{Journal of the American Statistical Association}, 67(337), 187--191.

\bibitem[{Yoon(2023)}]{5h7v-aq35-23}
Yoon, T. (2023).
\newblock Positional data of soft robots with diverse actuations types.
\newblock \doi{10.21227/5h7v-aq35}.
\newblock \urlprefix\url{https://dx.doi.org/10.21227/5h7v-aq35}.

\bibitem[{Yoon et~al.(2024)Yoon, Chai, Jang, Lee, Kim, Kwon, Kim, and Choi}]{yoon2024kinematics}
Yoon, T., Chai, Y., Jang, Y., Lee, H., Kim, J., Kwon, J., Kim, J., and Choi, S. (2024).
\newblock Kinematics-informed neural networks: Enhancing generalization performance of soft robot model identification.
\newblock \emph{IEEE Robotics and Automation Letters}, 9(4), 3068--3075.

\bibitem[{Zhou et~al.(2021)Zhou, Ren, Chen, Niu, Han, and Ren}]{zhou2021bio}
Zhou, L., Ren, L., Chen, Y., Niu, S., Han, Z., and Ren, L. (2021).
\newblock Bio-inspired soft grippers based on impactive gripping.
\newblock \emph{Advanced Science}, 8(9), 2002017.

\end{thebibliography}

\end{document}